\documentclass{sig-alternate}
\usepackage{graphicx,amsmath,amssymb,amsbsy,amsfonts,mathrsfs,amscd}

\usepackage{algorithm}
\usepackage{algorithmic}
\floatname{algorithm}{Procedure}

\usepackage{subfigure}
\usepackage[usenames]{color}
\def \g {\gamma}
\def \G {\Gamma}
\def \tg {\tilde{\gamma}}
\def \tG {\tilde{\Gamma}}

\def \reals {\mathbb{R}}

\def \E {\mathbb{E}}
\def \< {\langle}
\def \> {\rangle}
\def\P{\mathbb{P}}

\def\er{\mathcal{E}}
\def\SMM{{\sf SMM}}

\def\cons{\operatorname{cons}}
\def\simi{\operatorname{sim}}
\def\argmin{\operatorname{argmin}}
\def\argmax{\operatorname{argmax}}
\def\score{\operatorname{score}}
\def\normal{{\sf{N}}}
\long\def\hide#1{{}}

\DeclareMathAlphabet{\mathpzc}{OT1}{pzc}{m}{it}
\newtheorem{thm}{Theorem}[section]

\newtheorem{prepro}{{\bf Proposition}}[section]

\newtheorem{precor}{{\bf Corollary}}[section]

\newtheorem{preconj}{{\bf Conjecture}}

\newtheorem{preremark}{{\bf Remark}}[section]

\newtheorem{predef}{{\bf Definition}}

\newtheorem{prelem}{{\bf Lemma}}[section]

\newtheorem{preclaim}{{\bf Claim}}[section]

\begin{document}
%

\title{Multi-track Map Matching}
%
%
%
%
%

\numberofauthors{3} 
%
\author{
%
%
\alignauthor
Adel Javanmard\titlenote{This work was partially done while Adel Javanmard was an intern at Microsoft Research, Mountain View, CA.}\\
       \affaddr{Department of Electrical Engineering}\\
       \affaddr{Stanford University}\\
       \email{adelj@stanford.edu}
\alignauthor
Maya Haridasan\titlenote{Maya Haridasan was affiliated with Microsoft Research, Mountain View, CA at the time of this work.}\\
       \affaddr{Google, Inc.}\\
       \affaddr{Mountain View, CA}\\
       \email{haridasan@google.com}
\alignauthor Li Zhang\\
       \affaddr{Microsoft Research}\\
       \affaddr{Mountain View, CA}\\
       \email{lzha@microsoft.com}
}

\maketitle

\begin{abstract}
We study algorithms for matching user tracks, consisting of time-ordered location points, to paths in the road network. Previous work has focused on the scenario where the location data is linearly ordered and consists of fairly dense and regular samples. In this work, we consider the \emph{multi-track map matching}, where the location data comes from different trips on the same route, each with very sparse samples.  This captures the realistic scenario where users repeatedly travel on regular routes and samples are sparsely collected, either due to energy consumption constraints or because samples are only collected when the user actively uses a service. In the multi-track problem, the total set of combined locations is only partially ordered, rather than globally ordered as required by previous map-matching algorithms. We propose two methods, the iterative projection scheme and the graph Laplacian scheme, to solve the multi-track problem by using a single-track map-matching subroutine. We also propose a boosting technique which may be applied to either approach to improve the accuracy of the estimated paths. In addition, in order to deal with variable sampling rates in single-track map matching, we propose a method based on a particular regularized cost function that can be adapted for different sampling rates and measurement errors. We evaluate the effectiveness of our techniques for reconstructing tracks under several different configurations of sampling error and sampling rate.
\end{abstract}
%
%





\section{Introduction}

Map matching is the procedure for determining the path of a user on a
map from a sequence of location data (which we refer to as
\emph{track}). This process serves as a common preprocessing step for reasoning 
about traffic on the road network as well as for providing better location-based 
services~\cite{bpsw-05,klh-07,l-09,startrack-09,mit-11}. 
Converting a track to a topological path on a map
not only makes it easier to reason about paths, but
also leads to reduced storage requirements and more efficient 
operations on the same. Some examples of such operations are 
comparison of tracks, which is significantly more efficient to do on a discrete
road network than as polygonal lines on the plane, querying and
 retrieval of tracks based on similarity to other tracks~\cite{startrack-10}.

Due to its importance, many methods for map 
matching have been previously proposed~\cite{bpsw-05,klh-07,nk-09,l-09,mit-11,chen-11}, focusing on matching a single sufficiently dense and accurate
sequence of locations.  In this work, we consider
the \emph{multi-track map matching problem}, where we are given a
number of tracks generated from trips through the same path, and
 we wish to recover the underlying path that generates these tracks.  The problem is much more challenging than the single-track problem since each track contains a small number of samples (i.e. sampling intervals are large). This captures 
the realistic scenario where users repeatedly travel on regular routes,
and samples are sparsely collected due to restraints in energy consumption on the mobile device. 
Another scenario is when a service collects users' location information
only when the user actively uses such service. In such a scenario, the collected
location data can be very sparse. However, since users typically travel on the 
same (or similar) routes repeatedly, one may compensate the sparseness of 
the data by combining location data from different trips.

The main challenge in multi-track map matching is that in combining
data from multiple tracks, global ordering on all samples is not available,
a necessary condition for applying existing single-track map matching 
algorithms.  Instead, each track only provides the order
on a subset of locations.  If we apply the map matching algorithm on
each individual track, we would obtain paths with very poor quality
given the low sampling rate on each path. 

In this work we 
present multi-track  map matching algorithms in which partial orders
from input tracks are ``aggregated'' to form an appropriate global order, 
after which we can reduce the problem to single-track map matching. 


Given its role in our multi-track map matching approaches, 
we also revisit single-track map matching in order to understand
how to fine-tune it for scenarios when the measurement error is high,
and/or when the sampling rate of tracks is too high or too low. 
We study an approach to map-matching that involves minimizing
a regularized cost function including two types of errors in measuring the quality of a path: the data error
(fidelity of the path to the data points), and the model error (measuring the ``niceness" of the obtained path).  The two terms are balanced
according to a regularization parameter, and we study the optimal
choice of regularization parameter through both theoretical and
empirical studies.  We show that, for a simplified model, there is an
optimal choice of the regularization parameter. We also verify, through
experiments, that the algorithm behaves as predicted
by our theoretical study. 

In summary, our main contributions are
\begin{itemize}

\item We study the multi-track map matching problem, and 
propose two methods for solving it. Our general approach
 consists of (a) merging multiple tracks into a single one and (b) running
it through a single-track algorithm. We
 propose two methods for merging tracks: an iterative projection scheme and
a graph Laplacian scheme.  We also propose a generic framework to remove
 outliers by aggregating the matching results from multiple
 sub-samples.

\item We revisit the single-track map matching problem, formulate an
 optimization problem and prove rigorously that the solution to the
 problem achieves optimal path reconstruction in terms of the minimax
 risk for a simplified model.  While the optimization problem
 resembles previous hidden Markov model (HMM) methods, our approach
 allows a principled way to adapt its parameter according to properties 
of the input  data.

\item We evaluate the effectiveness of our proposed techniques for reconstructing tracks under several different configurations of sampling error and sampling rate. The evaluations are done on the dataset available in~\cite{seattleData}. The data set contains tracks collected from real users in Seattle, WA, using commercially available consumer grade GPS device. Our results indicate
that the proposed approaches lead to reasonable estimates of the route, significantly better than what would be achieved in case tracks were map matched individually.
\end{itemize}

%
\section{Related work}

Map matching has become an increasingly important problem over the past few years due to the proliferation of GPS tracking devices and track-based applications. A number of algorithms have been proposed to address this problem~\cite{nk-09,l-09,klh-07,mit-11,chen-11}.  A class of these works contain statistical methods which are based on Bayesian estimators. The authors in~\cite{l-09,nk-09} use the fact that the actual positions of the user on a path form a Markov chain. Given the location measurements, a hidden Markov model is defined with the actual positions on the path as the hidden states. The measurement probabilities in the model are determined based on the location measurement noise distribution (normal distribution with mean zero and variance $\sigma^2$); the transition probabilities are determined based on the spatial geometric and topological restrictions along with the temporal/speed constraints of the trajectories. The matched path is the one with maximum posterior probability. 

Another line of research does not use any statistical methods to address the problem. Authors in~\cite{chen-11} use curve simplification for approximating the Fr\'echet distance of curves. Given a polygonal curve $\pi$ and an embedded graph $G$ with edges embedded as straight line segments, they attack the problem of finding the closest path in $G$ to the curve $\pi$ with respect to the Fr\'echet distance. In~\cite{bpsw-05}, the average Fr\'echet distance is used to reduce the effect of outliers.

One of the contributions of our work is to introduce a graph Laplacian-based scheme for the multi-track map matching problem. Though graph Laplacians are widely used in machine learning for dimensionality reduction~\cite{Belkin03laplacianeigenmaps,diffusion}, spectral clustering~\cite{Luxburg,Ng01onspectral} and semi-supervised learning~\cite{Argyriou05combininggraph}, using them for map matching is a contribution of the present paper. Given the partial orderings on the locations, we construct an appropriate distance matrix and use the Laplacian of the corresponding weighted graph to find a global ordering of the locations. 
%

\section{Problem statement}

In this section, we define some notations and formally present the 
map matching problem. We assume that the user traverses a path $\Gamma$ on the
road network with some bounded velocity.  At time instants
$t_1,\cdots,t_n$, her location is recorded by the GPS device or
obtained by other localization methods. Each measured data consists
of a time-stamped latitude/longitude pair, which is subject to some noise. 
Denote the actual location of the user at time $t_j$ by $\g_j$ and 
let $\tg_j$ be the measured location at time $t_j$ ($\g_j, \tg_j \in \reals^2$).  

The location noise is distributed as a zero-mean Gaussian vector with variance
$\sigma^2$, i.e.,
\begin{equation*}
\g_j - \tg_j \sim \normal(0, 
\begin{bmatrix}
\sigma^2 & 0\\
0 & \sigma^2
\end{bmatrix}).
\end{equation*}

We call the time-stamped sequence $\tG = (\tg_1, \ldots, \tg_n)$ a
\emph{track}. Figure~\ref{fig:lambda_effect}(a) shows a portion of 
a path $\Gamma$, and (b) shows the corresponding track. In this paper, we 
consider the following two problems.

\begin{itemize}
\item \emph{Single-track map matching.} We are given a single track $\tG$. 
Since locations are time-stamped, there exists a global ordering on the 
locations in the time domain. The aim is to reconstruct the path $\Gamma$ from $\tG$.
\item \emph{Multi-track map matching.} Several user tracks $\tG_1,\cdots$, $\tG_m$  are available,
all generated from traveling on a single path $\Gamma$. This models the scenario where 
the tracks are collected over different days or from different users traveling 
across $\Gamma$. The goal here is to use all the tracks to recover the path. Note that in this case, 
there are only partial orders on the locations in the time domain. 
\end{itemize}

The map matching problem becomes more challenging when the location measurement 
error is high and/or when the sampling rate is too high or too low. Furthermore,
for the same number of sample points and the same measurement error,
the multi-track map matching is inherently more challenging as it lacks a global
ordering of the points.

In solving the map matching problem, we implicitly assume that the
user tends to travel on the shortest (quickest) path. This is an important
assumption that facilitates finding good matches.

In order to evaluate the quality of a map matching algorithm, we need a way to 
measure the similarity between two paths on the map.  We can view each path 
$\Gamma$ as the set of road segments it contains. We define the similarity between two paths as
\begin{equation}
\simi(\Gamma_1, \Gamma_2) = \frac{\ell(\Gamma_1 \cap \Gamma_2)}{\ell(\Gamma_1 \cup \Gamma_2)}\,,
\end{equation}
where for a set $S$ of road segments, $\ell(S)$ denotes the total
length of the road segments in $S$. Notice that this measure
captures both false negative and false positive segments 
on the matches.  In
particular, $\simi(\Gamma_1,\Gamma_2)\leq 1$ and
$\simi(\Gamma_1,\Gamma_2)=1$ iff both paths consist of the same set of
road segments. Our metric is slightly different from some previously used
metrics.  For example, in \cite{l-09}, the similarity measure ignores
false negatives which results in a more lenient metric than ours.

In the following, we denote the Euclidean distance by $\|\cdot\|$ and the shortest path distance by $\|\cdot\|_d$. For a path $P$ and two points $x,y\in P$, we denote the length of $P$ between the points $x,y$ by $\|xy\|_P$ .

\hide{
\begin{figure}[!h]
\centering
\includegraphics*[viewport = 100 50 480 750, width = 2in]{mapmatch1.pdf}
\caption{Map Matching Problem} \label{fig:mapmatch}
\end{figure}}

\section{Single-Track Map Matching}
\label{sec:smm}

\subsection{Algorithm}
Our method is based on minimizing a regularized cost function that balances two types of errors in measuring
the quality of a path: the \emph{data error}, which measures the
fidelity of the path to the data points, and the \emph{model error},
which measures the ``niceness'' of the obtained path.  

Formally, our method maps each $\tg_j$ to a point $x_j$ on the
road network. The produced path then consists of consecutive shortest
paths that connect $x_j$ and $x_{j+1}$ for $j=1, \ldots, n-1$.  For
the path defined by $X=(x_1,\ldots, x_n)$, the quality of the match is
measured by the following regularized cost function.
\begin{equation}\label{eqn:main1}
C_\lambda(\tG, X) = \sum_{j=1}^n \|x_j \tg_j\|^2 + \lambda \sum_{j=1}^{n-1} \|x_j x_{j+1}\|_d^2\,,
\end{equation}
where $\|xy\|_d$ represents the \emph{driving distance} between two
points $x$ and $y$ on the road network, i.e. the length of the
shortest path between $x$ and $y$.

The above cost function contains two terms: the former measures the
distance of the observed points to the path and corresponds to the data
error. The latter measures the local optimality of the path and
corresponds to the model error.  The regularization parameter
$\lambda$ balances between these two terms and plays a crucial role in
the estimator error.

Given the cost function $C_\lambda(\tG,X)$, the map matching algorithm
finds the sequence that minimizes $C_\lambda(\tG, X)$ to serve as the
matched path. We denote the outcome as 
\[A_\lambda(\tG) = \argmin_X C_\lambda(\tG, X)\,.\]

Since finding the global minimum is difficult, our implementation, to
be described later, actually finds an approximate solution.

The cost function $C_\lambda(\cdot,\cdot)$ is very similar to what
has been used in previous methods based on Hidden Markov 
Models (HMM)~\cite{klh-07,nk-09,l-09} (with minor variations in modeling
the error). However, in all previous work, $\lambda$ is set
to a constant.  One of the focus of our work is to study the impact of
$\lambda$ on the match quality and to present guidelines for choosing the
optimal $\lambda$ for a given input.

\subsection{Choosing the regularization parameter}

\begin{figure*}[ht]
\centering
\subfigure[Original Path]{
\includegraphics*[viewport = 0 100 650 665, width=2.5in]{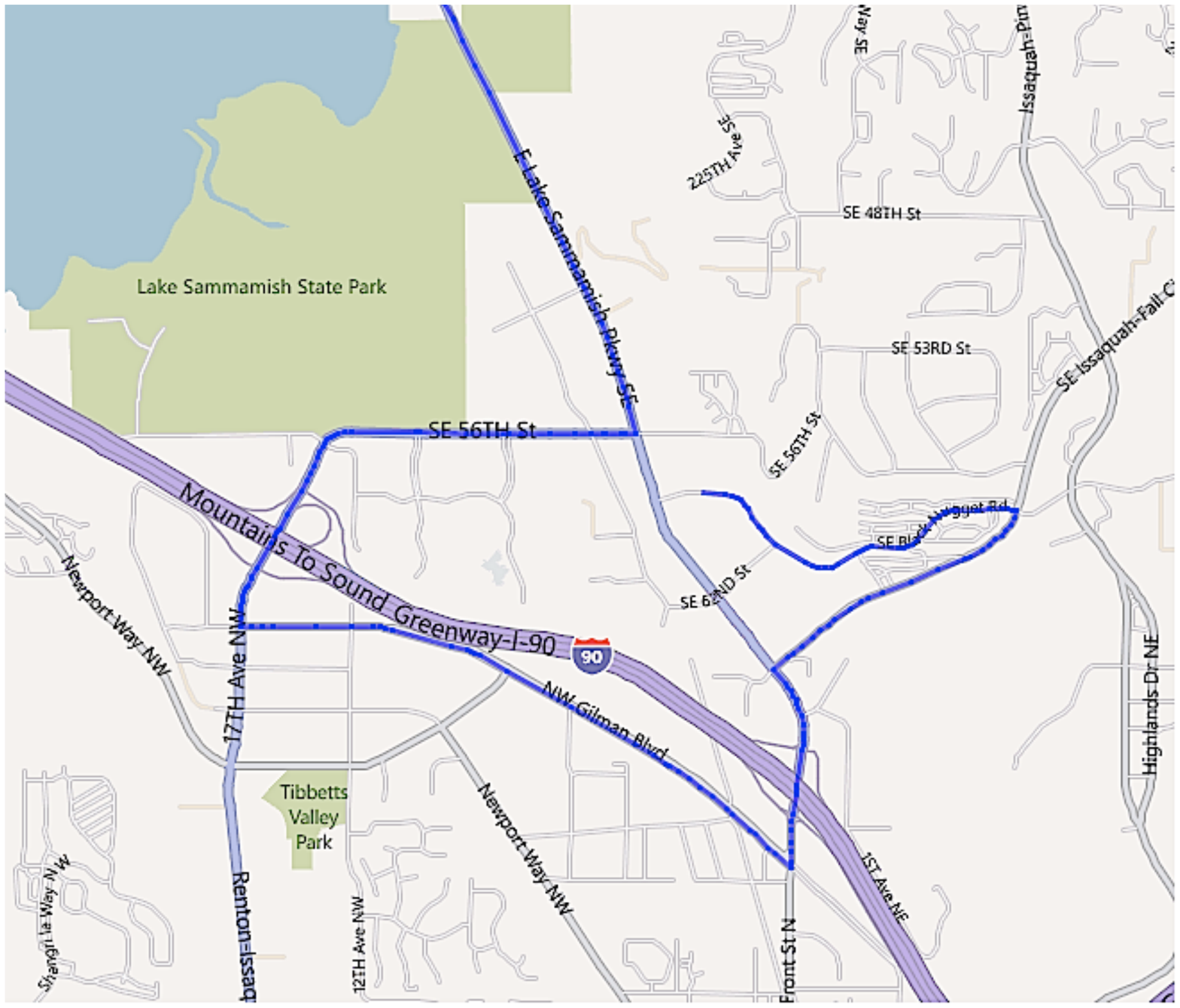}
}
\subfigure[Measured Locations]{
\includegraphics*[viewport = 0 100 650 665, width=2.5in]{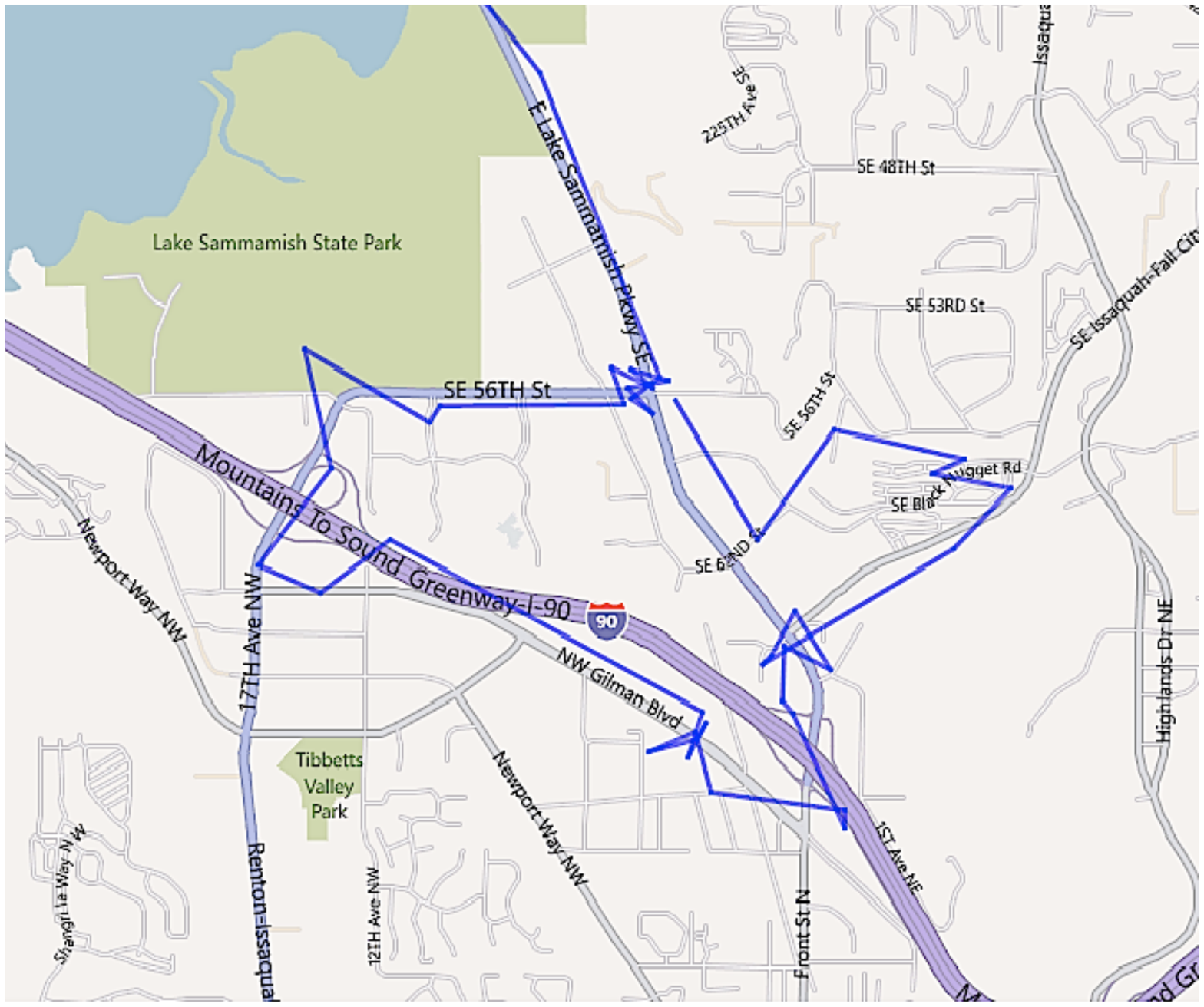}
}
\subfigure[Reconstructed Path, $\lambda = 10^2$]{
\includegraphics*[viewport = 0 100 650 665, width=2.5in]{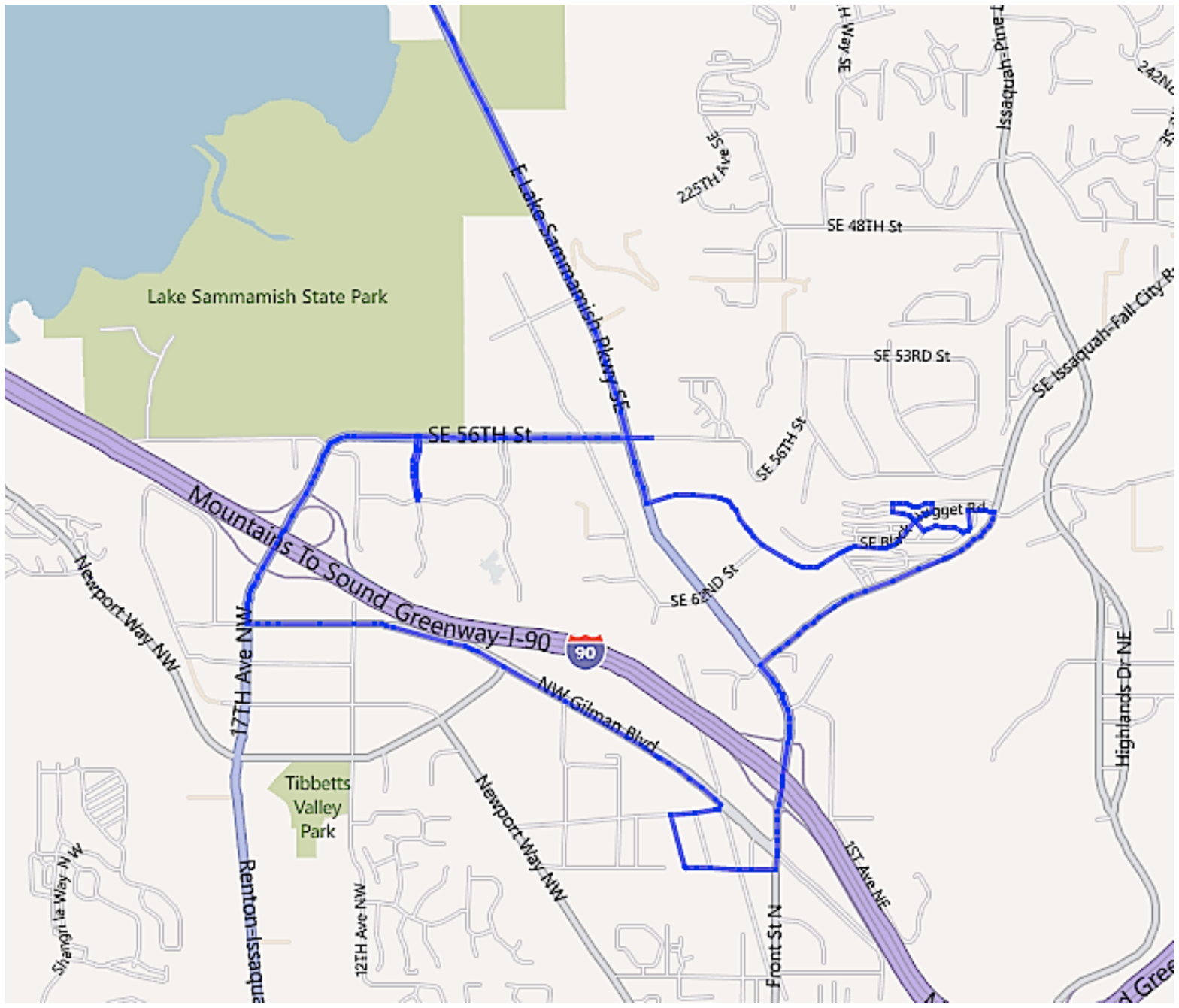}
}
\subfigure[Reconstructed Path, $\lambda = 10^{6}$]{
\includegraphics*[viewport = 0 100 650 665, width=2.5in]{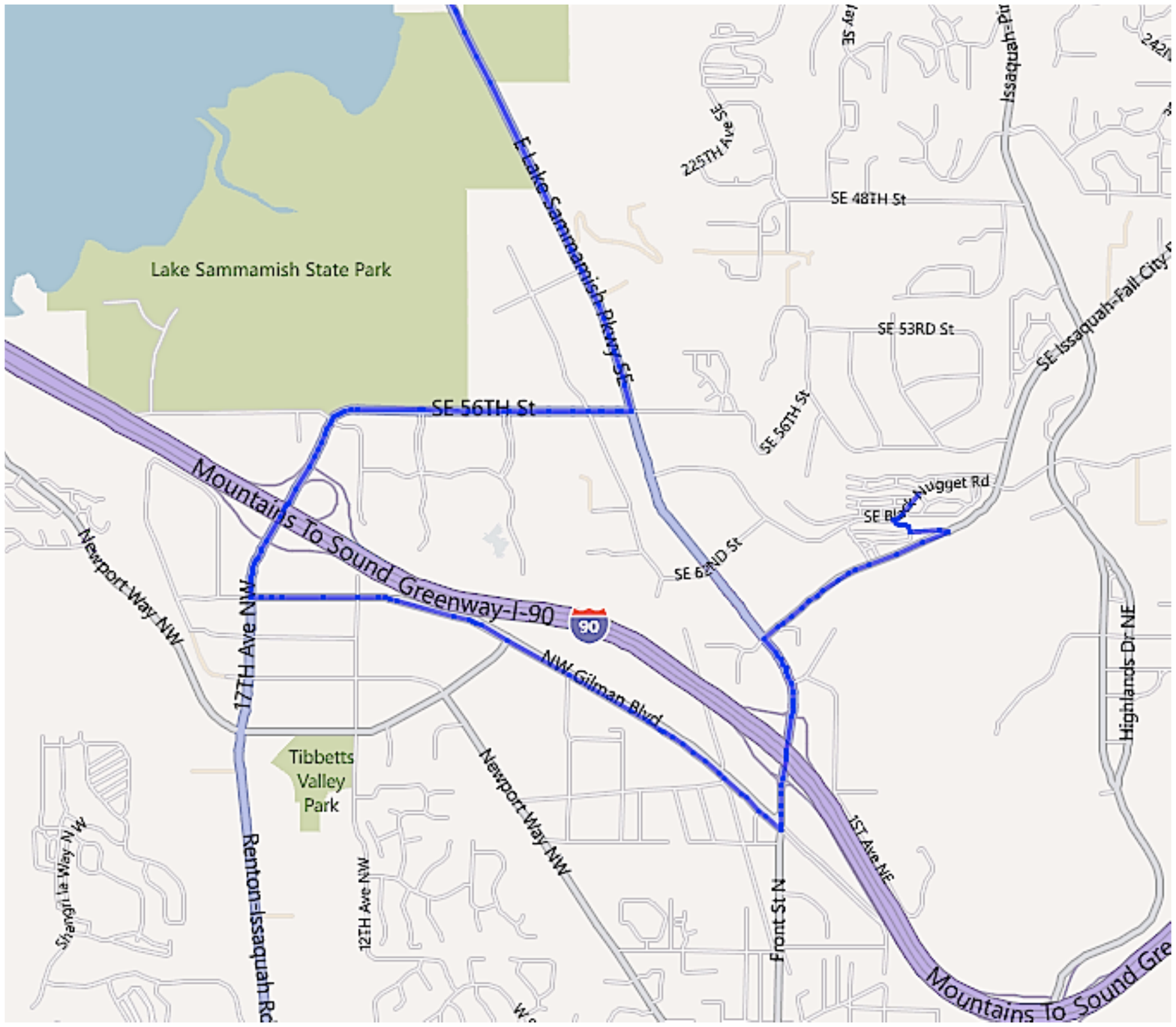}
}
\caption{\small The effect of regularization parameter $\lambda$. A small value of $\lambda$ results in strange loops, while a large value of $\lambda$ leads to missing geometrical details of the path.}
\label{fig:lambda_effect}
\end{figure*}

As discussed before, the regularization parameter $\lambda$ controls
the weight of the terms in the cost function. We will show, through
both theoretical and empirical studies, that the value of $\lambda$ has
significant impact on the matching quality. For the theoretical analysis, we
consider a simplified model and present an asymptotically optimal choice
of $\lambda$ for the model.  Later, we evaluate the choice of $\lambda$
through experiments with real data and show the same trend as 
theoretically predicted.

For an intuitive understanding of the effect of $\lambda$, consider
the extreme cases $\lambda = 0$ and $\lambda = +\infty$. Examples are
shown in Fig.~\ref{fig:lambda_effect}.
\begin{itemize}
\item {\boldmath{$\lambda = 0$. }} In this case, all weight is put on fidelity to the measured data. Therefore, the obtained path is the one passing through the projections\footnote{The projection of a point on a path is the nearest point on the path to the point.} of measured data onto the nearest road. While the location data is important as the sole indicator of the path, naively matching each noisy point to the nearest road will result in unsuitable paths involving short loops or U-turns, and overall a peculiar driving behavior.

\item {\boldmath{$\lambda = +\infty$. }} In this case, the emphasis is on finding a quick path; As a result, lots of geometrical details in the original path will be missing in the recovered one. 
\end{itemize}

For a better understanding of the effect of $\lambda$, we study a
simplified model and characterize the optimal choice of $\lambda$ for
it. In the model, we consider the situation where the sampling
rate is high so the shortest path distance can be approximated by the
Euclidean distance between the two endpoints. For a given
regularization parameter $\lambda$ and $\tG = (\tg_1,\ldots,\tg_n)$,
let $A'_{\lambda}(\tG)$ denote the optimal solution under $\lambda$,
i.e.
\begin{align*}
A'_{\lambda}(\tG) &=  \argmin_{X} C'_\lambda(\tG, X)\\
 & =  \argmin_{X} \sum_{j=1}^n \|x_j \tg_j\|^2 + \lambda \sum_{j=1}^{n-1} \|x_j x_{j+1}\|^2.
\end{align*}

Notice that in this model, we use the Euclidean distance rather than the
shortest path distance in the cost function.  To evaluate the quality
of matching, we adopt the standard minimax risk framework~\cite{dlm-90,johnstone-11}. Let
$\Gamma=(\gamma_1, \ldots, \gamma_n)$ be the ground truth. We require
$\Gamma$ to satisfy the condition 
\begin{equation}\label{eq:lipschitz} 
\|\gamma_j\gamma_{j+1}\|\leq c L/n\,,\;\;\;\;\;\; j =1 , \cdots, n-1\,,
\end{equation}
where $L$ is the total length of the path and $c$ is a constant
dependent on factors that upper-bound the distance between consecutive samples,
 such as speed-limit.  Recall that for an observed sequence of locations 
$\tG = (\tg_1, \cdots, \tg_n)$, we have $\tg_j =\gamma_j + \sigma g_j$, 
where $g_j$ are independent standard Gaussian noises. Therefore, $\tG$ follows
the distribution $\normal(\G, \sigma I)$. We
use the mean squared error to measure the quality of the output,
i.e. for a match $X=(x_1, \ldots, x_n)$, let $e(\G, X) =
\frac{1}{n}\sum_j \|x_j\gamma_j\|^2$ denote the error of $X$. 
Then the expected error of any map matching algorithm $M$ under the
ground truth $\G$ is
\begin{equation}
\er(M,\G) = \E_{\tG \sim \normal(\G, \sigma I)} e(\G, M(\tG))\,.
\end{equation}

The minimax risk of a map matching method is $R(M)=\max_\Gamma
\er(M,\Gamma)$, where $\Gamma$ is taken among all the samples
satisfying (\ref{eq:lipschitz}). We can show that
\begin{thm}\label{thm:main1}
With the above notation, we have 
\begin{equation}\label{eq:err}
R(A'_\lambda) \leq  c_1 \lambda\left(\frac{L}{n}\right)^2 + c_2 \frac{\sigma^2}{\sqrt{\lambda}}\,,
\end{equation}
for some constant $c_1,c_2>0$. $R$ is minimized when
$\lambda^\ast=\Theta((n\sigma/L)^{4/3})$ with $R(A'_{\lambda^\ast}) =
O((\sigma^2L/n)^{2/3})$. Furthermore, for any estimator $M$, $R(M) =
\Omega((\sigma^2L/n)^{2/3})$.
\end{thm}

The proof of the above theorem requires some involved analysis which
we omit from this abstract. Instead, we discuss some implications of
the above theorem.

$1)$ Large values of $\lambda$ put emphasis on minimizing the
 path length and may lead to large data errors (first term in
 (\ref{eq:err})). Meanwhile, small values of $\lambda$ allow the 
model to become finely tuned to noise, potentially leading to 
large model errors (second term in (\ref{eq:err})).  The optimum 
choice of $\lambda$ is a balance of the two error terms.

$2)$ The optimal $\lambda^\ast = \Theta((n\sigma/L)^{4/3})$
 increases with either $n$ or $\sigma$ and decreases with
 $L$. Intuitively, this means that we need to put more weight on the
 model error when the location measurement is noisy or when the sampling rate
 is high.  We shall verify this observation later via experiments.

 $3)$ The optimal minimax risk, $O((\sigma^2L/n)^{2/3}) \to 0$ when
 $n\to \infty$.  In other words, noise and under sampling have a
 similar effect on the error. For any value of $\sigma^2$, the error
 can be made arbitrarily small by increasing the sampling rate. But this
 is not true if we choose a constant $\lambda$ since the second term
then remains constant.  This partly explains the paradoxical phenomenon, as
 observed in \cite{nk-09}, in which the quality of a map matching
 algorithm deteriorates when the sampling rate is very high.

 $4)$ The algorithm based on the regularized cost function is
 asymptotically optimal (in the minimax risk order of magnitude) among all the
 matching algorithms! We find this quite surprising given the simple
 formulation of the algorithm and the vast options of estimators.

Theorem~\ref{thm:main1} and the above discussion applies to 
the simplified model in which we replace the shortest path distance by the Euclidean
distance.  However, as we shall show in our experiments, the above
statements qualitatively hold for the map-matching problem. Thus they
serve as good guidelines for choosing $\lambda$.
  
\subsection{Estimating $\sigma$}

As shown above, the optimal choice of $\lambda$ depends on the measurement
noise $\sigma$ and the average distance between consecutive samples $L/n$. While the latter
can be easily found through the data, we need to estimate $\sigma$
from the input.  We use a cross validation technique to estimate the
parameter $\sigma$. We divide the samples into a disjoint training
set and a validation set.  We then use the map-matching algorithm 
on the training set to obtain a path $P$.  Next, we compute the distance 
from each point in the validation set to $P$, and use the average distance 
to determine $\sigma$.  

To achieve better accuracy, we need to make sure that the training set
has sufficient samples to produce a high quality match.  This is
controlled by taking a sparse subset of regularly spaced samples as
the validation set.  A more formal description can be found in
Procedure~\ref{algo:cross}. (for a path $\g^{(i)}$ and a location $\tg_k$, the notation
$\mathcal{P}_{\g^{(i)}}(\tg_k)$ in Procedure~\ref{algo:cross} denotes the projection of $\tg_k$ onto the path $\g^{(i)}$).
\begin{algorithm}[!h]
\caption{Cross Validation for estimating $\sigma$}\label{algo:cross}
\begin{algorithmic}[1]
\REQUIRE Measured locations $\{\tg_k\}$ for $k=1,\ldots,n$.
\ENSURE The estimation $\hat{\sigma}$.
\STATE Set $m = \lfloor \sqrt{n} \rfloor$.
\FOR {$i \in \{1, \dots, m\}$}
\STATE Let $S_i = \{\tg_{i}, \tg_{m+i}, \tg_{2m+i}, \dots\}$;
\STATE Match a path $\g^{(i)}$ to the points in $S_i^c$;
\STATE Let 
\[d_i = \frac{1}{|S_i|} \sum_{\tg_k \in S_i} \|\tg_k \mathcal{P}_{\g^{(i)}}(\tg_k)\|^2\,;\]
\STATE Set $\hat{\sigma}_{i} = \sqrt{d_i/2}$;
\ENDFOR
\STATE Return $\hat{\sigma} = 1/m \sum_{i=1}^m \hat{\sigma}_{i}$.
\end{algorithmic}
\end{algorithm}

One remaining question is the choice of $\lambda$ in the map matching
algorithm in Procedure~\ref{algo:cross} (step 4).  As shown in the experiments, for a fairly large range of
$\lambda$, the reconstruction accuracy is still adequate to estimate
$\sigma$.

\subsection{Implementation}

\begin{figure*}[ht]
\centering
\subfigure[Matched Candidates for measurement $\tg_j$]{
\includegraphics*[viewport = 0 50 600 500, width=2.5in]{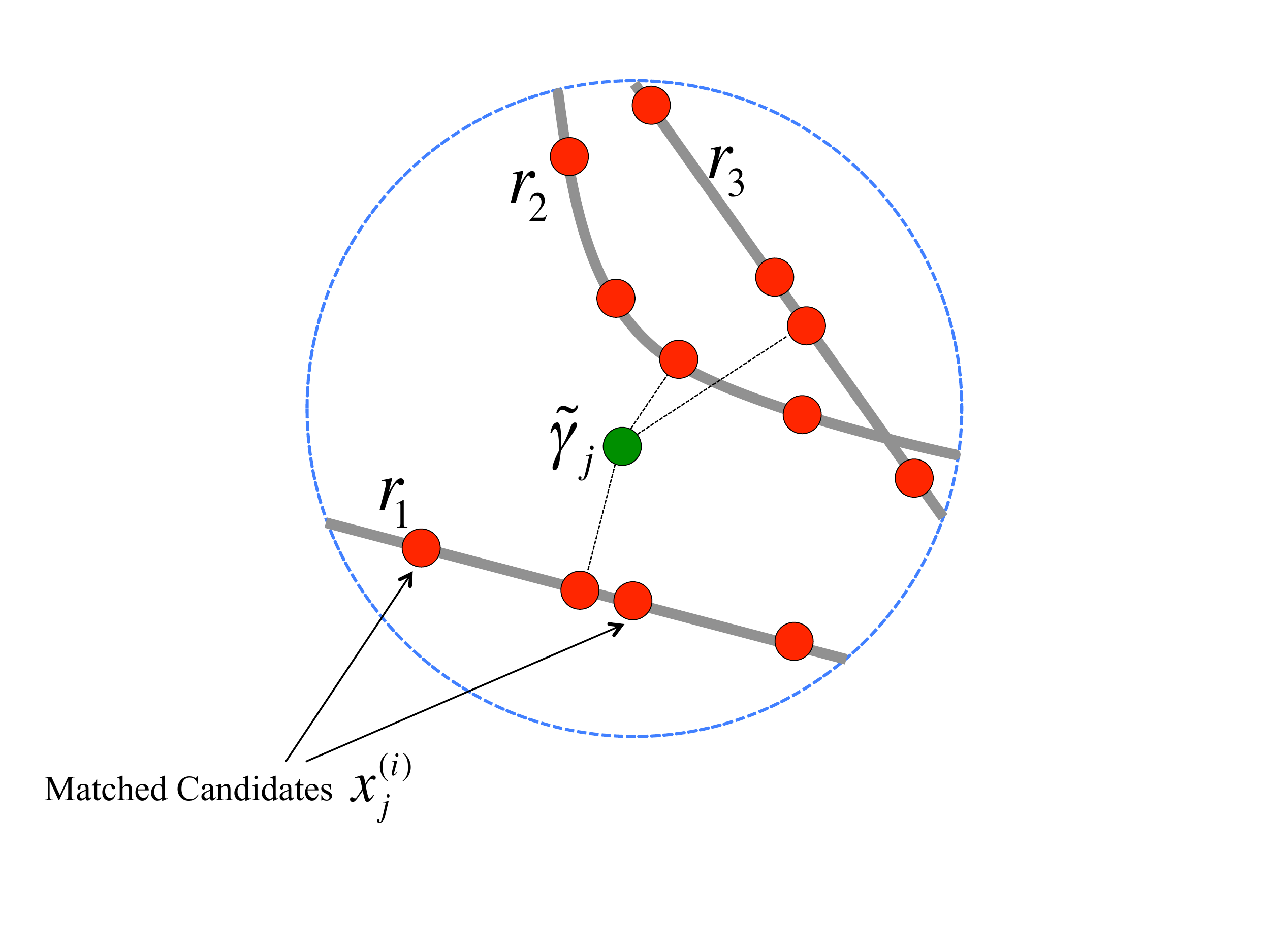}
}
\hspace{0cm}
\subfigure[Graph of matched candidates]{
\includegraphics*[viewport = 0 50 720 500, width=3.5in]{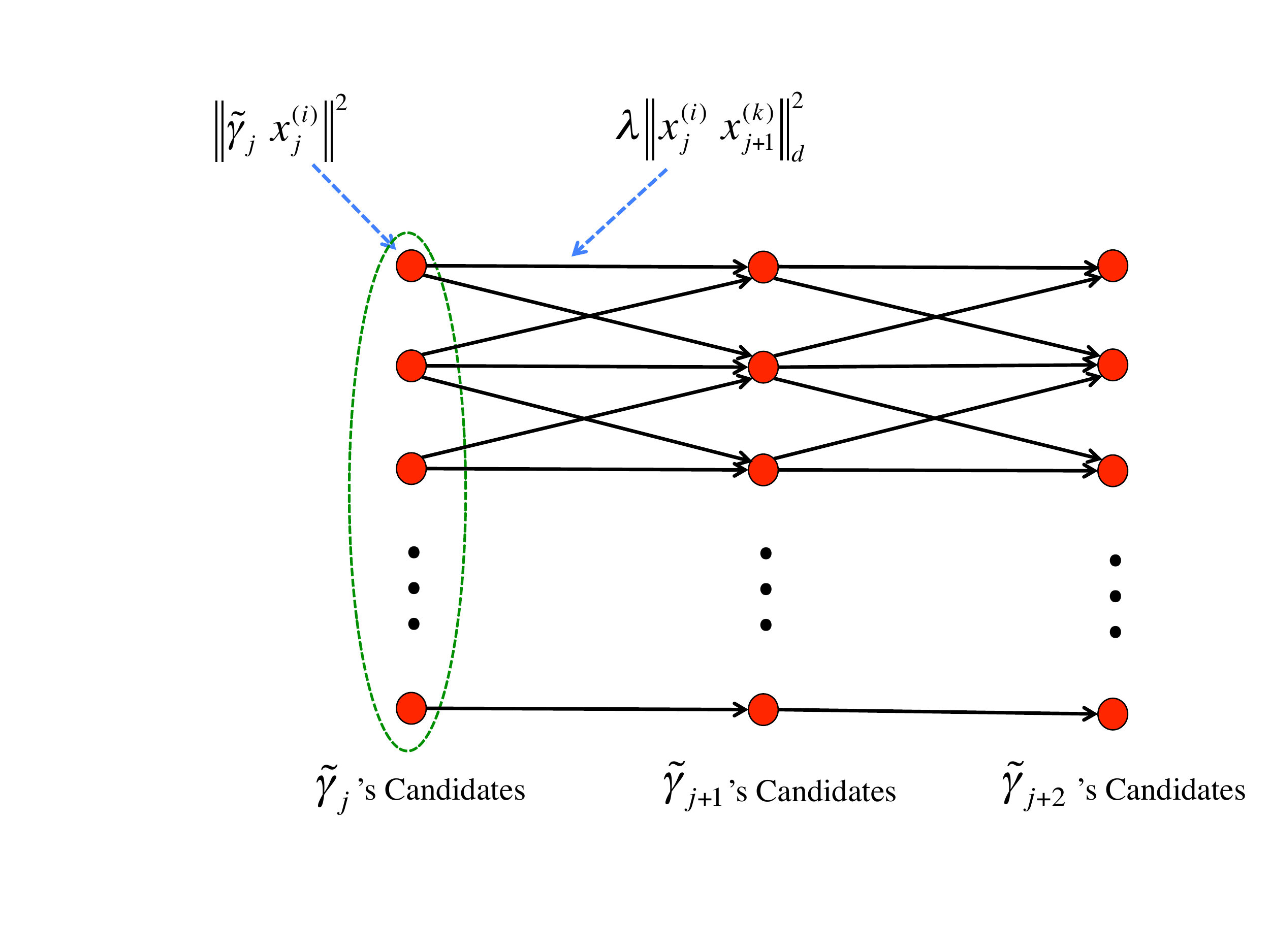}
}
\caption{{\small  We use dynamic programming to find the minimum weighted path in the graph of matched candidates. In Fig (a), $N=3$, hence we have $4$ matched candidates on each road segment $r_1,r_2, r_3$.}}
\label{fig:matchcandidate}
\end{figure*}

The global optimization with respect to the cost function~\eqref{eqn:main1}
is possible but time consuming.  We employ a
pruning procedure to reduce the solution space to a smaller set of
candidates and then apply dynamic programming to compute
the minimum solution in the pruned set.  The resulting algorithm is
similar to HMM-based methods~\cite{klh-07,nk-09,l-09}. 

We first construct a multipartite graph $G$ with the $j^{th}$ part
corresponding to the measurement $\tg_j$.  For each part, we consider
a small set of road points, which we call \emph{match candidates}, one
of which will be matched to $\tg_j$.  The match candidates should be
chosen such that they represent well all the possible points that
might have generated $\tg_j$. On the other hand, we need to keep its
size small for fast computation.

In our implementation, for each $\tg_j$, we consider the road segments
that are within $200$ meters from it. 
On each of these road segments, we consider $N+1$ match candidates including the nearest point
on the road segment to $\tg_j$ along with $N$ other points evenly spaced on that road segment. (see Fig.~\ref{fig:matchcandidate} for an illustration). 
The role of these $N$ extra candidates (per segment) is
to increase the algorithm flexibility in choosing matches,
especially when the location error is large or when the sampling rate
is too high. In all of our experiments, we choose $N=3$.  To every
vertex in part $j$, corresponding to point $x$ on the map, we assign
the value $\|\tg_j x\|^2$. 

We then connect every vertex in part $j$ to every vertex in part
$j+1$.  To each edge between points $x$ and $y$, we assign the edge
weight $\lambda\|xy\|_d^2$.  We then compute the minimum
weighted path in graph $G$ and output the points on the path as the
match.  The weight of the path is calculated by summing up all the edge
and the vertex weights on the path. We use standard dynamic
programming algorithm to compute the optimal path in this graph.  To
efficiently compute the edge weights, we use the
\emph{contraction hierarchy} shortest path software developed
by~\cite{gssd-08}.

\section{Multi-Track Map Matching}

In the single track map matching problem, the global ordering on the
locations is known a priori since timestamps are available for each
location. In the Multi-track map matching problem, ordering of
locations is also available for each individual track. However, it is
not known how samples from different tracks are ordered with respect
to each other when we merge them into a single track.  In this
section, we propose techniques that take samples from multiple tracks
refering to the same route and combine them into a single track with
total ordering. Once total ordering is achieved, single-track map
matching may be applied to the combined track.  We first propose two
methods to obtain a global ordering on locations. We then introduce a
\emph{boosting process} that further enhances the performance of both
methods.

Before proceeding to the methods, we establish some notations. For a
set of measured locations $S$ and an order $\pi$ on $S$, we denote by
$\SMM(S,\pi)$ the outcome of the single track map matching algorithm
described in Section~\ref{sec:smm}.  When $S$ consists of a single
track, we simply write it as $\SMM(S)$ as the order is already given
in the track.

\subsection{Iterative projection method}

For any given path $P$, we define the order of a set of points $S$
with respect to $P$ as follows. We first compute, for each point $s\in
S$, the nearest point $\tilde{s}$, called the projection of $s$, on
$P$. We then order $S$ according to the order of $\tilde{s}$ on $P$.
In the iterative projection method, we choose an initial path and then
order all the points with respect to the path. Once we obtain the
order, we run the single-track map matching algorithm on the points
with the computed order. The resulting path becomes the new candidate
path, and we repeat the above process until either the process
converges or it has run for too many rounds. The algorithm is
summarized in Procedure~\ref{algo:iterative}.  

The quality of the algorithm depends on the choice of the initial
path. At the extreme, if the initial path is just a point, then the
projection method would project all the points to the same point and
thus fails to produce any meaningful result.  A good initial path
should ``cover'' all the points, i.e. pass through most
points. However, we do not require the initial path to have very
precise geometry as it will be corrected nicely during the iteration.
With these constraints, we choose the starting track as the one that
minimizes the sum of distances from samples in all the other tracks to it, 
where the distance from a sample to a track is the shortest
distance between the sample and the track. This often leads to a good
candidate, but occasionally we end up with bad initial path. The
\emph{boosting process} that we later describe helps alleviate this
problem.

In order to quickly compute the nearest point, we implemented a
quad-tree~\cite{quad-tree} data structure so that for any given path
represented by a polygonal line, we can compute efficiently, for any
query point, its nearest point on the path.

\begin{algorithm}
\caption{Iterative projection method}\label{algo:iterative}
\begin{algorithmic}[1]
\REQUIRE a set of GPS locations $S$ from tracks $\tG_1,\cdots,\tG_k$.
\ENSURE global ordering $\pi$ on $S$.
\STATE Choose the initial track $S_0$ from $S_1,\cdots, S_k$ 
\STATE Set $P = \SMM(S_0)$;
\REPEAT
\STATE Compute the order $\pi$ of $S$ with respect to $P$;
\STATE Set $P = \SMM(S,\pi)$.
\UNTIL{($P$ remains the same) or (too many rounds)}
\STATE return $\pi$
\end{algorithmic}
\end{algorithm}

\subsection{Graph Laplacian Method}

If we ignore the order of the GPS locations, the map-matching problem
resembles the classical curve reconstruction problem in which one is
asked to reconstruct a curve from discrete samples on the curve. In
our second approach, we apply graph Laplacian method, an effective
method in machine
learning~\cite{Belkin03laplacianeigenmaps,diffusion,Ng01onspectral,
Argyriou05combininggraph}, to compute the order of the GPS locations
and then apply the single track map matching on the resulted sequence
of GPS locations.  We will first describe the standard graph Laplacian
method and then make modification to adapt to the particular problem
of map matching.

\medskip

\noindent\textbf{Graph Laplacian method.}
Graph Laplacian works in any metric space. Suppose that we are given
$n$ points and the $n\times n$ distance matrix $D$. The goal is to
find an order of the $n$ points that is consistent with
$D$. Intuitively, an order is consistent if nearby points in the order
are also close-by according to $D$.  The graph Laplacian method
computes such an order by first solving the following minimization problem
\begin{eqnarray}
\begin{split}
\underset{v}{\text{minimize}} & \quad \sum_{1\le i \le j\le n} (v_i - v_j)^2/D_{ij}\label{eq:lap}\\
\mbox{subject to} &\quad \quad \|v\| = 1,\\
&\quad \quad \sum_j v_j=0\,,
\end{split}
\end{eqnarray}
where $\|v\| = \sqrt{\sum_{i} v_i^2}$. 
Given the solution $v^\ast$ of (\ref{eq:lap}), we sort the values
of $v^\ast_j$ for $j=1,\ldots, n$ to obtain an ordering $\pi$ of the
points.  The intuition of the above approach is quite clear: if $i,j$
are close, i.e. $D_{ij}$ is small, we would then like $|v_i-v_j|$
to be small in order to minimize the objective function, which in turn
implies that $i,j$ would be close in the resulted ordering.  The
constraints are to avoid the trivial solution of setting all $v_j$'s to some constant or scaling $v_j$'s to very small number.

The optimization problem (\ref{eq:lap}) can be efficiently solved by
computing the eigenvectors of Laplacian matrix $L$ defined as
\[ L_{ij} = \left\{
\begin{array}{ll}
-1/D_{ij} & i\neq j\\
\sum_{k\neq i} 1/D_{ik} & i=j
\end{array}
\right.
\]

This is because $v^T L v = \sum_j(v_i-v_j)^2/D_{ij}$. Therefore, $L$ is
positive semi-definite and the minimum eigenvalue of $L$ is $0$
corresponding to the eigenvector $(1, \ldots, 1)^T$.  Since the
eigenvectors of any symmetric matrix are orthogonal to each other, the
eigenvector corresponding to the second smallest eigenvalue is exactly
the solution to (\ref{eq:lap}).

The quality of the Laplacian method depends crucially on the accuracy
of $D$ as an approximation of the distances along the curve.  If it is
exact, then the Laplacian method produces the exact answer.  One
practical heuristic is to replace $D_{ij}$ by $\exp(c D_{ij})$ so to
put more weights on the short distances as they are more trustworthy
approximation to the distances along the curve.  In the following, we will
describe how we form the distance matrix in order to more faithfully
approximate the shortest path distance between the points.

\medskip

\begin{figure}[ht]
\includegraphics*[viewport = 20 280 750 450, width = 3.5in]{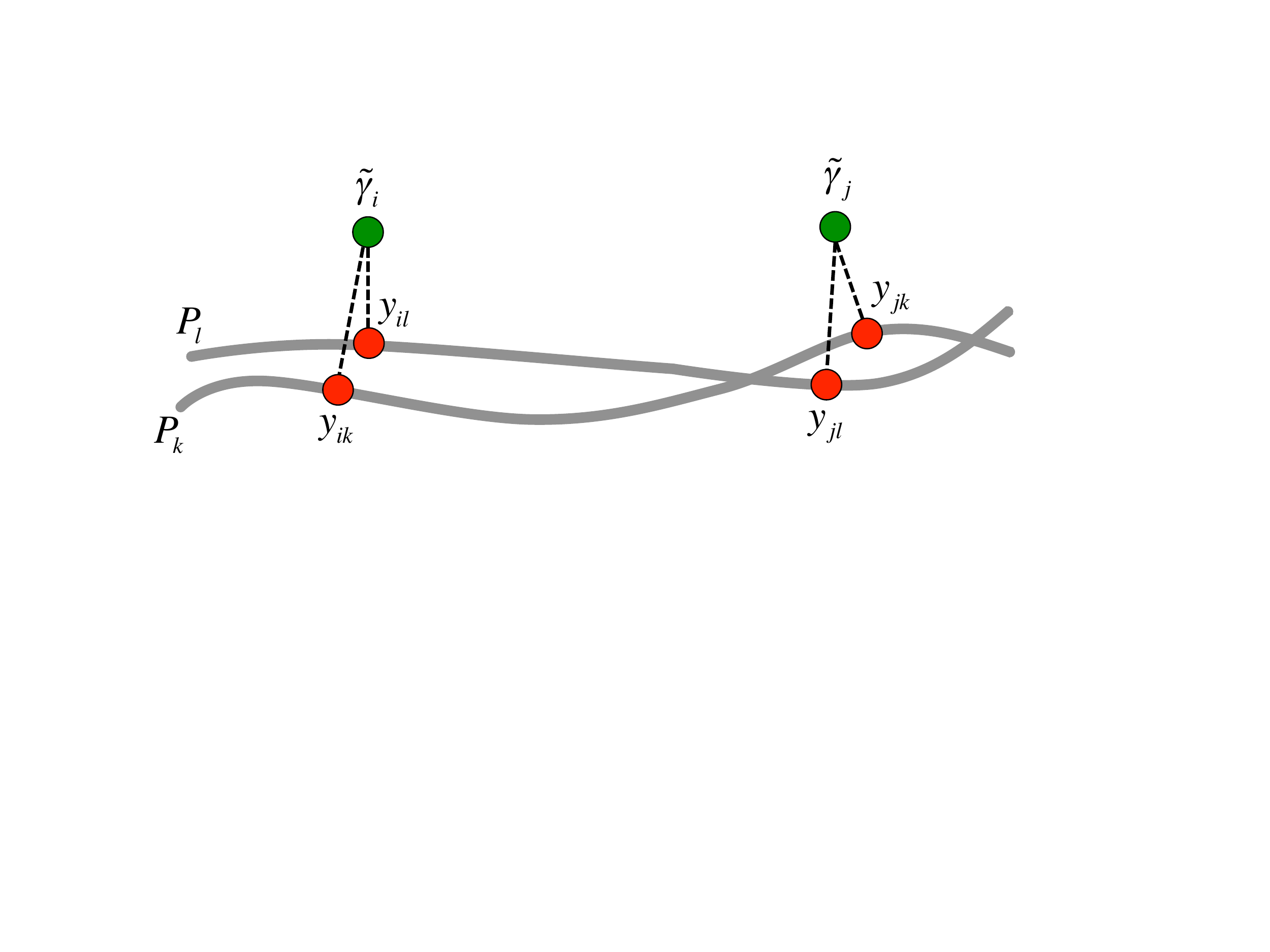}
\caption{{\small Computing distance matrix. $y_{ik},y_{il}$ ($y_{jk},y_{jl}$) are the closest points to $\tg_i$ ($\tg_j$) on the path $P_k$ and $P_l$, respectively. 
$d^{(k)}_{ij}=\|\tg_i y_{ik}\|+\|\tg_j y_{jk}\| + \|y_{ik}y_{jk}\|_{P_k}$,
$d^{(l)}_{ij}=\|\tg_i y_{il}\|+\|\tg_j y_{jl}\| + \|y_{il}y_{jl}\|_{P_l}$, and $D_{ij} = (d^{(k)}_{ij}+d^{(l)}_{ij})/2$.}}\label{fig:dist}
\end{figure}

\noindent\textbf{Constructing $D$ for map matching.}
The straight forward solution would be to use the Euclidean distance
between points to form the distance matrix.  However, this
approximation can be quite poor, especially for points far away from
each other or for points which are close together in the space but far
away by travel (think of two points on the opposite banks of a river).
To remedy these problems, we utilize the order information coming from
each individual track. Basically, when we compute the distance between
two sample points on different tracks, we use the distance along the
matched path to each individual track.  Figure~\ref{fig:dist} shows an
example on how the distance is computed.

Procedure~\ref{proc:lap} describes the details of how we compute the
distance matrix. Recall that $\|xy\|_P$ denotes the distance along the
path $P$ for two points $x,y\in P$.
\begin{algorithm}
\caption{Form the Laplacian matrix}\label{proc:lap}
\begin{algorithmic}[1]
\REQUIRE $n$ points $\tg_1, \ldots, \tg_n$ from $m$ tracks $\tG_1, \ldots, \tG_m$
\ENSURE  $D$: an $n\times n$ distance matrix
\FOR{$i \in \{1, \ldots, m\}$}
\STATE $P_i = \SMM(\tG_i)$
\ENDFOR
\FOR{$i \in \{1, \ldots, n\}$}
\FOR{$j \in \{i+1, \ldots, n\}$}
\STATE Suppose $\tg_i \in \tG_k$ and $\tg_j \in \tG_l$;
\STATE Let $y_{ik},y_{il}$ be the nearest point to $\tg_i$ on path $P_k$ and $P_l$, respectively. (see Fig.~\ref{fig:dist})
\STATE Let $y_{jk},y_{jl}$ be the nearest point to $\tg_j$ on path $P_k$ and $P_l$, respectively.
\STATE $d^{(k)}_{ij} =  \|\tg_i y_{ik}\| +  \|\tg_j y_{jk}\| + \|y_{ik}y_{jk}\|_{P_k}$;
\STATE $d^{(l)}_{ij} =  \|\tg_i y_{il}\| +  \|\tg_j y_{jl}\| + \|y_{il}y_{jl}\|_{P_l}$;
\STATE $D_{ij}= (d^{(k)}_{ij} + d^{(l)}_{ij})/2$.
\ENDFOR
\ENDFOR
\end{algorithmic}
\end{algorithm}

We then form the Laplacian by using the exponential weight,
\[ L_{ij} = \left\{
\begin{array}{ll}
-\exp(-c D_{ij}) & i\neq j\\
\sum_{k\neq i} \exp(-c D_{ik}) & i=j
\end{array}
\right.
\]

Once we have $L$, we just invoke the standard Laplacian method,
i.e. compute the eigenvector $v$ corresponding to the second
smallest eigenvalue of $L$; sort the values in $v$; return
the sorted order $\pi$ as the global order of the points.

\subsection{Boosting Process}

\begin{figure}[ht]
\centering
\includegraphics*[viewport = 20 80 750 420, width = 3.3in]{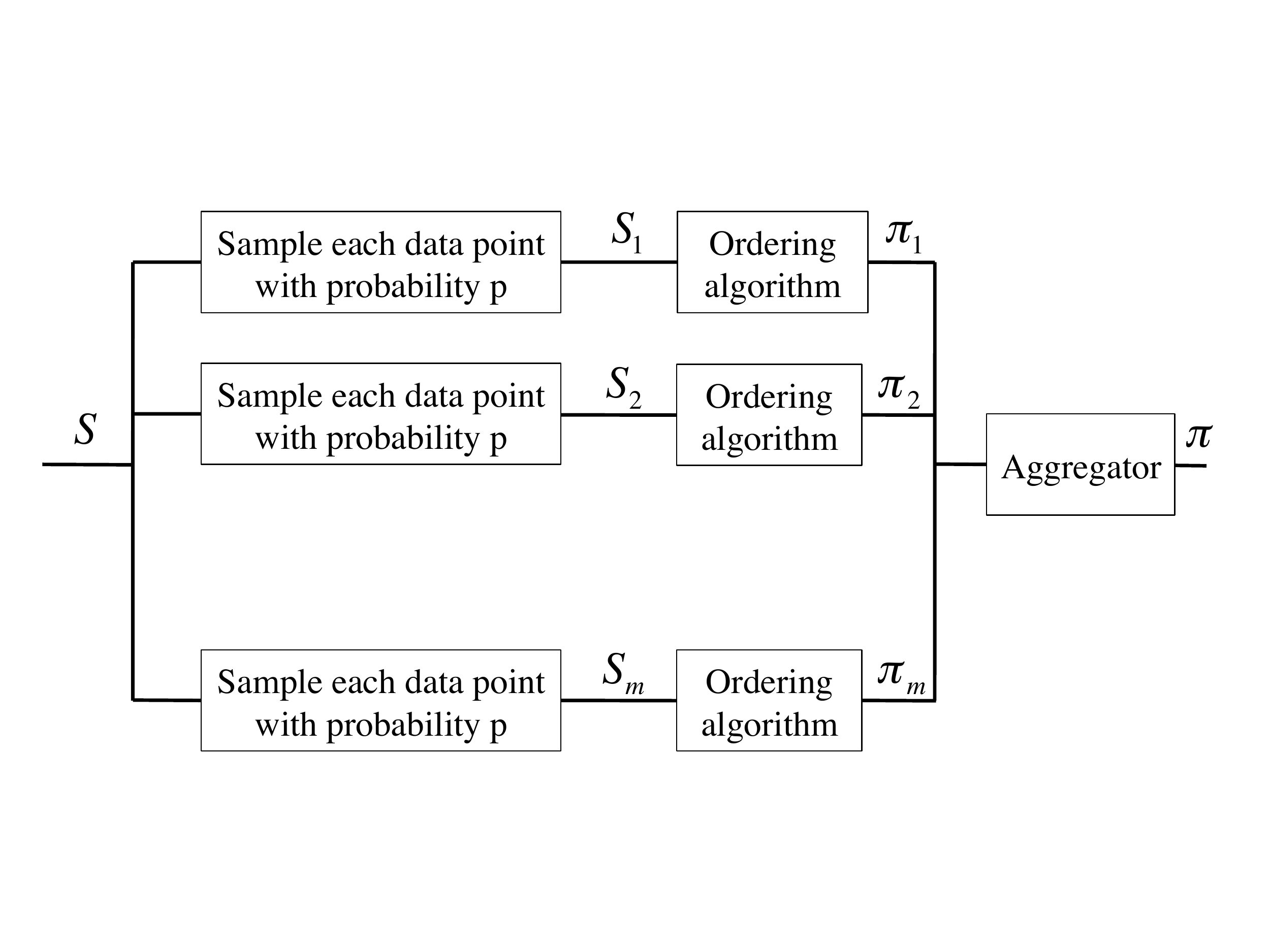}
\caption{\small Scheme for the boosting process.} \label{fig:boosting}
\end{figure}

Both the iterative projection scheme and the Laplacian method are
susceptible to noises, especially when there are outlier points with
large error.  The outlier points may cause either method to get trapped
into some wrong path.  To fix this problem, we introduce a boosting
process to improve the robustness of both methods against outliers.
Fig.~\ref{fig:boosting} shows the boosting scheme.

The boosting process is fairly generic and does not depend on any
particular ordering method.  From the set $S$, we generate $m$
subsets, namely $S_1,\dots,S_m$. Each $S_i$ is obtained by sampling
the data points in $S$ with some probability $p$. Hence $\E|S_i| = p
|S|$. Then, an ordering algorithm (either the iterative or the
Laplacian method) is used to return a global ordering on the set
$S_i$, denoted by $\pi_i$. The
\emph{aggregation} block in the algorithm takes the orderings
$\pi_1,\dots,\pi_m$ as input and returns an ordering $\pi$ which is
the most consistent one to the orderings $\pi_1,\cdots,\pi_m$ in the
following sense.

The \emph{consistency score} of two orders is defined as the
number of pairs they agree on subtracted by the number of
those they do not.  Here we do not require the two orders to be defined on
the exactly same set of elements. When they do not, we only consider
the elements common to both. Formally, 
\begin{align*}
\cons(\pi_1,\pi_2)&= |\{(i,j):\; (\pi_1(i)-\pi_1(j))(\pi_2(i)-\pi_2(j))>0\}|\\
&-|\{(i,j):\; (\pi_1(i)-\pi_1(j))(\pi_2(i)-\pi_2(j))<0\}|.
\end{align*}
The aggregator then attempts to find
\[\pi^* = \underset{\pi \in \Pi}{\argmax} \sum_{k=1}^m \cons(\pi, \pi_k)\,,\]
where $\Pi$ is the set of all possible orderings on all subsets of
$S$.  The reason that we do not insist $\pi$ to be a full order is
that we would allow the flexibility of excluding outliers from the
input data.

Unfortunately, computing the most consistent order of a set of
permutations is NP-hard.  Our aggregator block uses heuristics to find the
solution.  We first create a directed acyclic graph
(DAG) on the points which are consistent with the set of permutations.
We then compute the longest path in the DAG and output it as the
aggregated order.  The details are shown in Procedure~\ref{proc:agg}.
\begin{algorithm}[ht]
\caption{Order aggregation.}\label{proc:agg}
\begin{algorithmic}[1]
\REQUIRE A set of linear ordering $\pi_1, \ldots, \pi_m$
\ENSURE A consistent order $\pi^\ast$ 
\STATE  Let the scoring matrix $M \in \reals^{n \times n}$ be defined by
\[M_{ij} = |\{\pi_k: \pi_k(i) < \pi_k(j)\}| - |\{\pi_k: \pi_k(i) > \pi_k(j)\}|\,.\]
\STATE  For any permutation $\pi$, define 
\[\score(\pi) = \sum_{i,j} M_{ij} 1_{\{\pi(i) < \pi(j)\}}\,;\]
\STATE $\pi_M = \phi$;
\FOR{$k\in \{1, \ldots, K\}$}\label{state:local}
\STATE Pick a random permutation $\pi$;
\WHILE{$\exists\;i,j\quad \score(\texttt{swap}(\pi,i,j))>\score(\pi)$}
\STATE $\pi = \texttt{swap}(\pi, i, j)$;
\ENDWHILE
\IF{$\score(\pi)>\score(\pi_M)$}
\STATE $\pi_M=\pi$;
\ENDIF
\ENDFOR\label{state:local-end}
\STATE Form $G=(V,E)$ where $(i,j)\in E$ iff $M_{ij}>0$ and $\pi_M(i)<\pi_M(j)$; assign weight $M_{ij}$ to the edge $(i,j)$;
\STATE Compute the maximum weighted directed path $P$ in $G$;\label{state:lp}
\STATE Let $\pi^\ast$ be the order determined by the path $P$.
\end{algorithmic}
\end{algorithm}

In the description $\texttt{swap}(\pi, i,j )$ returns a permutation by
swapping the elements in the $i$-th and $j$-th positions in $\pi$.  In
the lines 4-12, we repeatedly pick
a random permutation as a starting point and find a locally most
consistent ordering through local searches.  We repeat the process $K$ times with a given $K$. In our experiments, we find it is sufficient to set $K$ to $100$.   We then keep the best order among $K$ trials.
Each round
finishes in $O(m n^2)$ iterations as each swap increases the score by
at least $1$, and the maximum score is bounded by $O(m n^2)$.  In
practice, the convergence is much faster, usually within $O(n)$
iterations.  Line~14 can be done in linear time since $G$ is
a directed acyclic graph.

\section{Experiments}
\label{sec:experiment}

We evaluate the performance of the proposed algorithms on data generated from real
tracks. We will first describe the data generation process and then present the experimental results.

\subsection{Data}

We utilize dataset of tracks collected from real users in Seattle,
WA~\cite{seattleData}. These data are collected using commercially
available consumer grade GPS device and are sampled with high
density. To test the map matching algorithms, we generate synthetic
data from these tracks with any given measurement error and sampling rate.

More specifically, from the dataset, we remove the paths that are too
short or too long.  Table~\ref{table:data} summarizes some statistics
of the paths we use.
\begin{table}[ht]
\caption{Statistics of data}\label{table:data}

\medskip

\begin{tabular} {l|r|r|r}
\hline
& Length (km) & Duration (min) & Segments\\
\hline
mean & 14.5 & 19 & 77 \\
\hline
minimum & 3.4 & 6 & 20 \\
\hline
maximum & 38.4 & 37 & 169\\
\hline
\end{tabular}
\end{table}

For a given path, we first sample in the time domain and determine at
which time to take a sample. We then use a randomly generated speed
on each road segment to sample the location on the path. Then we add
the location error to produce a sample location.
Procedure~\ref{algo:sampling} describes how we sample from a given
path with a given location error $\sigma$ and sampling interval $\tau$.

\begin{algorithm}[ht]
\caption{Generating sample from a path}\label{algo:sampling}
\begin{algorithmic}[1]
\REQUIRE a path $P$, measurement error $\sigma$, sampling distribution and rate $\tau$
\ENSURE a sequence $\tg_1, \tg_2, \ldots$

\STATE Generate time steps $t_1, t_2, \ldots$ according to the required sampling distribution and rate $\tau$;
\STATE Associate with each edge $e$ in $P$ a speed $v_e$ sampled uniformly at random from $[0.8 V_e, 1.2 V_e]$ where $V_e$ is the speed limit on the edge $e$;
\STATE Using the speed, compute the location $\gamma_j$ associated with each $t_j$ on path $P$;
\STATE Let $\tg_j=\gamma_j + \sigma g$ where $g$ is a standard Gaussian;
\STATE Produce $\tg_1, \tg_2, \ldots$ with associated time stamps $t_1, t_2, \ldots$. 
\end{algorithmic}
\end{algorithm}

For the sampling distribution, we experiment with both the uniform and
the exponential distributions. Under the uniform distribution, we
generate time-steps regularly at $\tau, 2\tau$, $
\cdots$; under the exponential distribution, we generate a series of
times where the inter-arrival time between any two adjacent samples
follows the exponential distribution with mean $\tau$.  The
exponential distribution better models the location data obtained when
the user uses location-based services.

\subsection{Single-track map matching}

\begin{figure*}[t]
\centering
\subfigure[]{
\includegraphics*[viewport = 10 10 250 180, width = 3in]{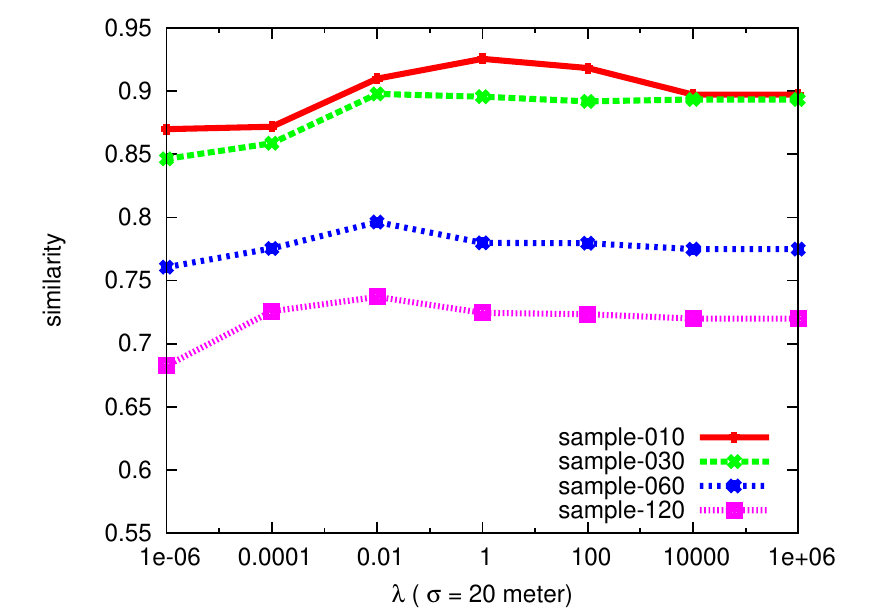}
}
\subfigure[]{
\includegraphics*[viewport = 10 10 250 180, width = 3in]{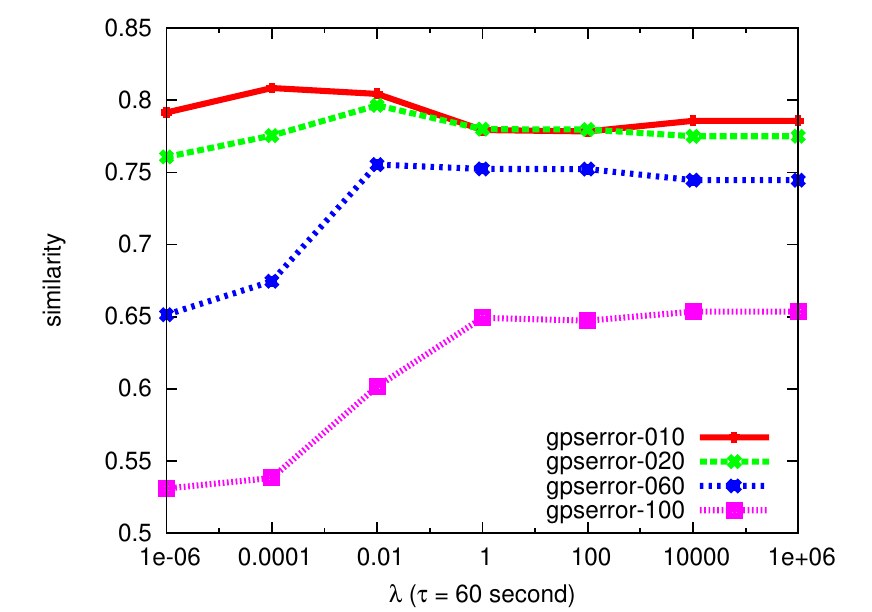}
}
\caption{{\small Similarity of single-track map matching algorithm. (a) $\sigma=20$ meters, varying $\tau$ and $\lambda$; (b) $\tau=60$ seconds, varying $\sigma$ and $\lambda$.}}
\label{fig:single}
\end{figure*}

In the first experiment, we run the single track map matching
algorithm on the first $20$ tracks in the data set and for various
combinations of $\sigma, \tau$ and $\lambda$.  For each track and each
combination of parameters, we generate $10$ instances and take the
average of the similarity of the middle $8$ results.  We then compute
the average similarity over $20$ tracks for each combination of
$\sigma, \tau$ and $\lambda$.  The results for $\sigma=20$ and
$\tau=60$ are shown in Fig.~\ref{fig:single}.

From these experiments, we can make the following observations.

\begin{enumerate}

 \item The similarity result is quite stable, but the choice of
 $\lambda$ has a visible impact on the quality.  For example, for
 $\sigma=20$ and $\tau=10$ (the topmost curve in
 Figure~\ref{fig:single}(a)), there is a gap of $8\%$ between the
 result when $\lambda=1$ and $\lambda=10^{-6}$.

 \item The experiments show similar trends as predicted by
 Theorem~\ref{thm:main1}.  From Figure~\ref{fig:single}(a), we can see
 that with $\sigma=20$ fixed, when the sampling rate increases, i.e. $n$
 decreases, the optimal $\lambda$ decreases, shifting from $1$ to
 $0.01$. From Figure~\ref{fig:single}(b), when fix $\tau=60$, optimal
 $\lambda$ increases when $\sigma$ increases too.
\end{enumerate}

\subsection{Multi-track map matching}

We now evaluate the performance of the different techniques proposed for multi-track map matching, i.e., the \emph{iterative} and the \emph{Laplacian} methods and their boosted variants. In the following experiments, tracks are generated using an exponential distribution. We collect results for 25 real routes taken by users, and for each route, generate a number $s$ of track instances.  (Different instances can be thought of as measurements collected over different days).

\begin{figure}[!t]
\centering
\includegraphics*[viewport = 10 10 250 180, width = 3in]{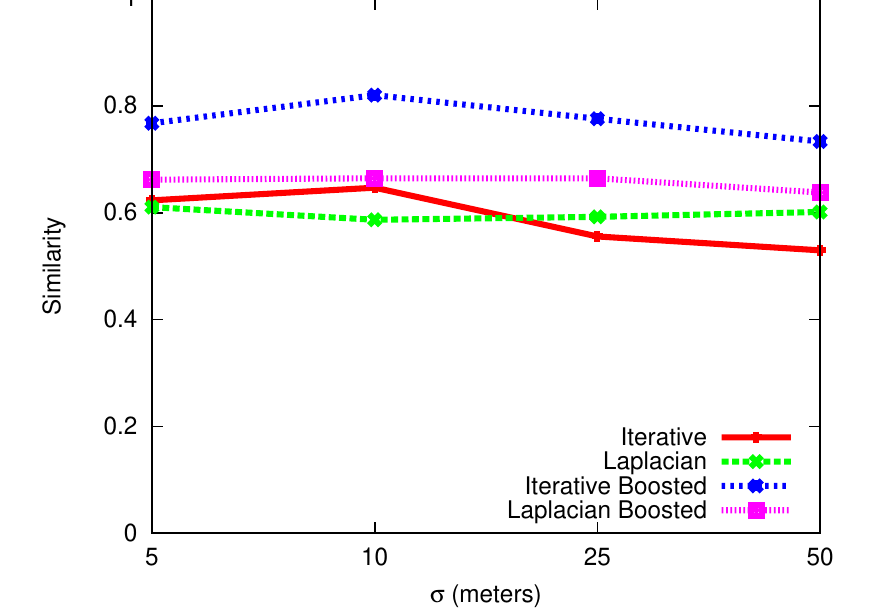}
\caption{{\small Average similarity when computing multi-track map matching with $s=20$ tracks. Sampling period of tracks is $T = 300 s$, and the measurement error $\sigma$ is varied from 10 to 200 meters.}} 
\label{fig:multi_comp}
\end{figure}

First, we compare the four methods for computing a multi-track map matched route from $s = 20$ track instances, synthetically generated with a sampling period $T = 300 s$ and with variable measurement errors. For the boosting process, we choose $m=10$ and $p=0.5$, i.e. we run the algorithm for $10$ subsamples and in each run include each point with probability $0.5$. Therefore, a point is included in five sub-samples on average.  We compute the similarity of the outcome relative to the ground truth. The average results when running the algorithm over the 25 different routes are presented in Fig.~\ref{fig:multi_comp}. We removed the top and the bottom $10\%$ similarity results prior to computing the average.

From the figure, we observe that in the non-boosted versions, the iterative and the Laplacian methods produce similar results, independent of the measurement error. Meanwhile, the boosted variants of both approaches improve the performance of their equivalent non-boosted versions. As shown in Fig.~\ref{fig:multi_comp}, the boosted iterative method returns better matches when the measurement error is smaller, but the quality of the results deteriorates faster as the measurement error increases. The boosted variant of the Laplacian method outperforms all other methods once the error is larger than 100m. This happens because in the iterative method, each point is mapped to the nearest neighbor on another path. This makes it more sensitive to the location measurement errors.

\begin{figure}[!t]
\centering
\includegraphics*[viewport = 10 10 250 180, width = 3in]{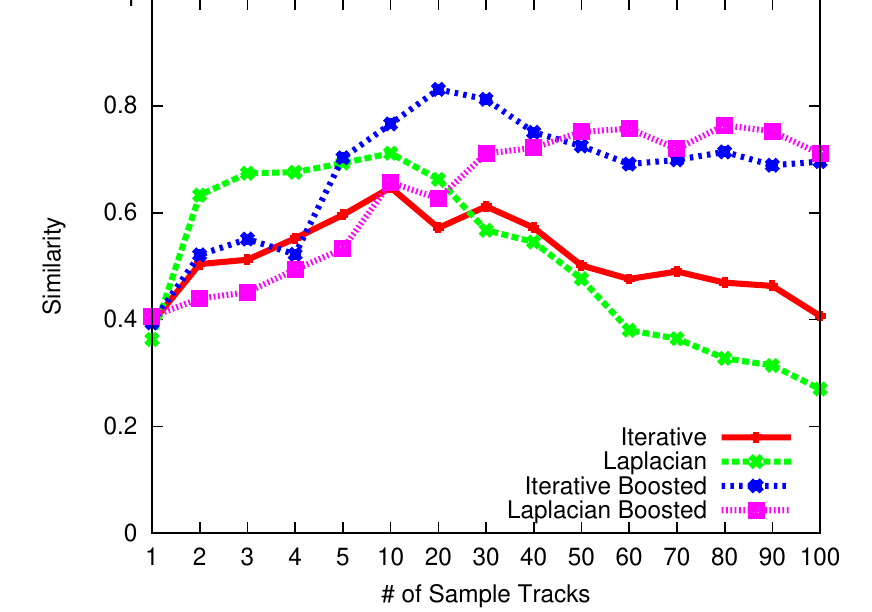}
\caption{{\small Average similarity when computing multi-track map matching with increasing numbers of tracks ($s=1$ to $100$). Sampling period of tracks is $T = 300 s$, and the measurement error is $\sigma=10m$.}} 
\label{fig:multi_comp_sampleCount}
\end{figure}

We also studied the behavior of the methods as the number of available
sample tracks per route increases. In
Fig.~\ref{fig:multi_comp_sampleCount} we vary the number of
sampled tracks from 1 to 100 and present the average similarity between
the obtained routes and the ground truth. In this experiment we fixed
the measurement error to $\sigma = 10 m$ and the sampling period to $T =
300s$. The results show the boosted versions significantly
outperforming the non-boosted variants once the number of samples is
large enough (5 for the iterative method and 20 for the Laplacian method). Once the
number of samples is larger than 45, the boosted Laplacian method starts
outperforming the boosted iterative method. This happens because 
the Laplacian method is more robust than the iterative method, and therefore
the effect of boosting is only visible when the number of tracks is large.

\begin{figure*}[!t]
\centering
\subfigure[Iterative]{
\includegraphics*[width = 3in]{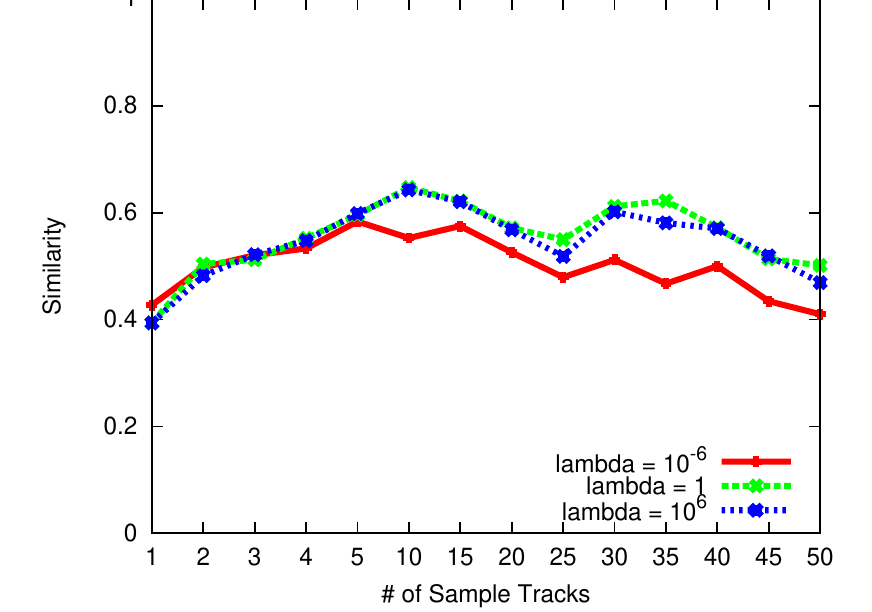}
}
\hspace{0cm}
\subfigure[Laplacian]{
\includegraphics*[width = 3in]{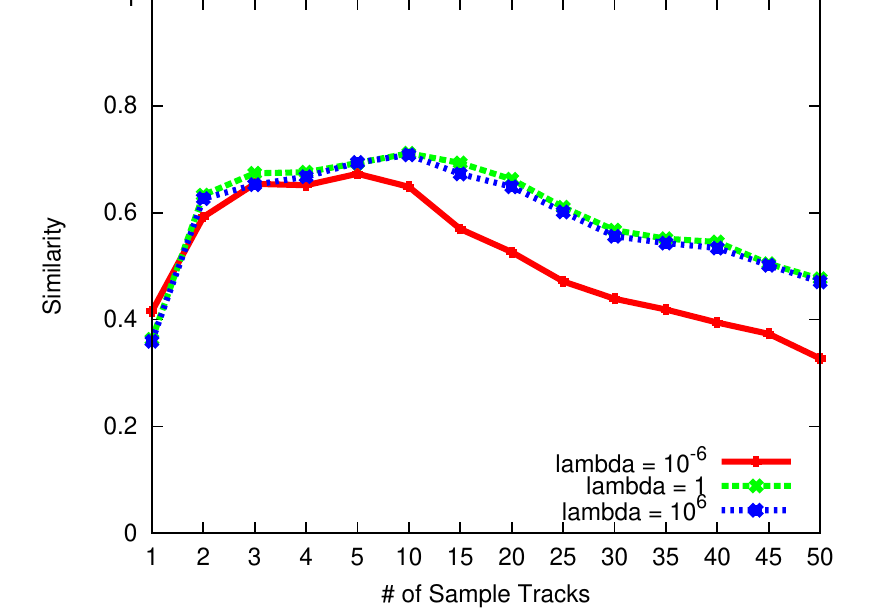}
}
\subfigure[Iterative Boosted]{
\includegraphics*[width = 3in]{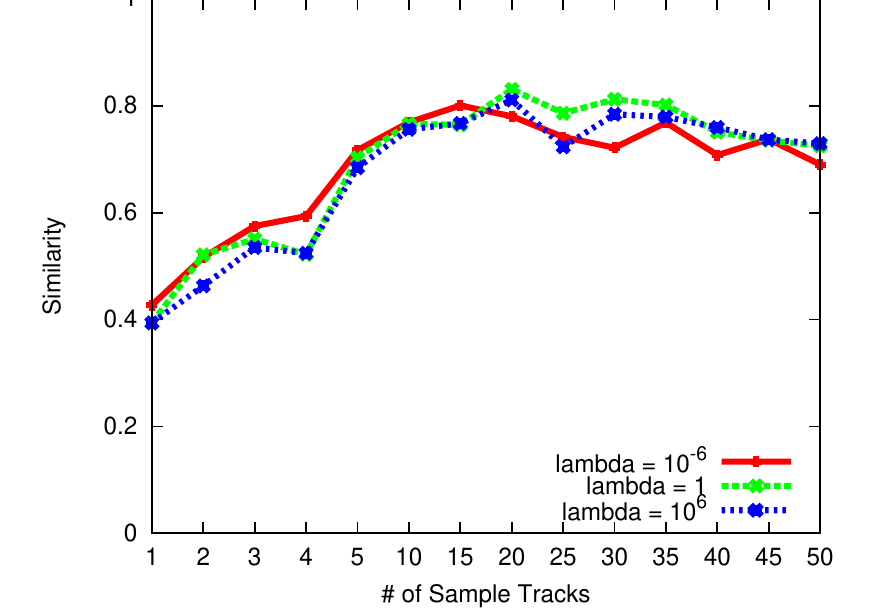}
}
\hspace{0cm}
\subfigure[Laplacian Boosted]{
\includegraphics*[width = 3in]{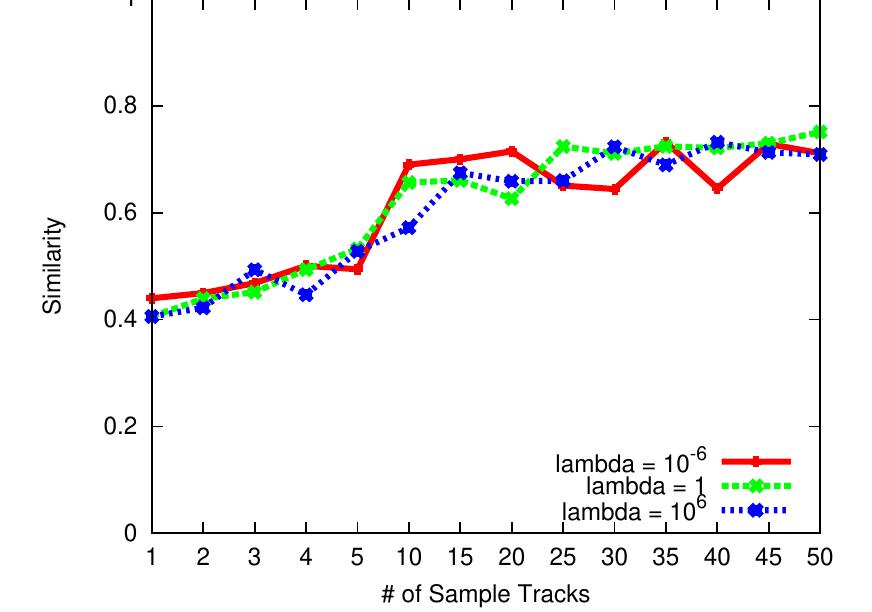}
}
\caption{{\small Average similarity when computing multi-track map matching with increasing numbers of tracks ($s=1$ to $50$) tracks. Sampling period of tracks is $T = 300 s$, and the measurement error is $\sigma=10m$.}}
\label{fig:lambda_effects}
\end{figure*}

Finally, we studied the effect of the regularization parameter
$\lambda$ used in the single-track subroutine when computing
multi-track map matching with these four
techniques. In Fig.~\ref{fig:lambda_effects} we present the average
similarity values when lambda is $10^{-6}$, $1$ and $10^6$, for all four techniques,
as we increase the number of track instances $s$ from 1 to 50. As
observed, the non-boosted variants are more affected by the choice of
lambda than their boosted counterparts. This happens because in the boosted
version, the sub-sampling process effectively reduces the density of
the samples and therefore makes the results less sensitive to the
change of $\lambda$.

\section{Conclusion}

In this paper, we defined and studied the multi-track map matching problem, 
in which multiple very sparse tracks of location samples on a single route are 
combined together and used to recover the underlying route. We tackled this problem by breaking it into a 
two step process: We first computed a global ordering on the entire set of 
samples; once a global ordering on the samples is computed, the problem reduces to the single-track 
map matching. In the second step, we solved the corresponding single-track map matching problem. 

We proposed and evaluated two algorithms for obtaining a 
global ordering on the samples, namely the \emph{iterative projection approach} and the \emph{graph Laplacian} 
approach, as well as a boosting process that helps remove 
outliers in the samples. Our results indicate that the proposed approaches 
lead to reasonable estimates of the route, significantly better than what would
be achieved in case tracks were map matched individually. 

We also revisited the single-track map matching problem, which is an essential
building block for the multi-track variant. We formulated the single-track map matching problem as a regularized optimization 
problem where the regularization parameter can be fine-tuned based on the properties of the map matched data. 
We evaluated the effect of the regularization parameter under different configurations
of measurement error and sampling rates, and showed how the parameter
may be successfully varied to achieve best results under different settings.

\section{Acknowledgments}
We thank Paul Newson, John Krumm, and Eric Horvitz for providing us with the experimental data and for their map-matching package. We thank Daniel Delling, Renato Werneck, and Andrew Goldberg for their shortest path software and very helpful discussions.

%

\bibliographystyle{abbrv}
\bibliography{mapmatching}  
\end{document}